\documentclass[conference]{IEEEtran}
\IEEEoverridecommandlockouts

\usepackage{cite}
\usepackage{amsmath,amssymb,amsfonts}
\usepackage{algorithmic}
\usepackage{graphicx}
\usepackage{subcaption}
\usepackage{booktabs}
\usepackage{textcomp}
\usepackage{xcolor}
\usepackage{url}
\usepackage{hyperref}
\usepackage{cleveref}
\usepackage{xcolor}

\def\BibTeX{{\rm B\kern-.05em{\sc i\kern-.025em b}\kern-.08em
T\kern-.1667em\lower.7ex\hbox{E}\kern-.125emX}}

\begin{document}

\title{Augmentations for Robust and Efficient Imitation Learning in Streamed Video Games}

\author{
\IEEEauthorblockN{
Somjit Nath\textsuperscript{1\dag},
Abdelhak Lemkhenter\textsuperscript{2},
Pallavi Choudhury\textsuperscript{2},
Chris Lovett\textsuperscript{2}
}
\IEEEauthorblockN{
Katja Hofmann\textsuperscript{2},
Sergio Valcarcel Macua\textsuperscript{2},
Lukas Sch\"afer\textsuperscript{2}
}
\IEEEauthorblockA{
\textsuperscript{1}McGill University, Mila \textsuperscript{\dag}Work done while at Microsoft. \quad
\textsuperscript{2}Microsoft
}
\thanks{\copyright~2026 IEEE. Personal use of this material is permitted. Permission from IEEE must be obtained for all other uses, in any current or future media, including reprinting/republishing this material for advertising or promotional purposes, creating new collective works, for resale or redistribution to servers or lists, or reuse of any copyrighted component of this work in other works.}
}

\IEEEoverridecommandlockouts
\IEEEpubid{\makebox[\columnwidth]{979-8-3315-9476-3/26/\$31.00~\copyright2026 IEEE \hfill}
\hspace{\columnsep}\makebox[\columnwidth]{}}

\maketitle
\IEEEpubidadjcol

\begin{abstract}
Imitation learning is an appealing way to scale game-playing agents to complex 3D environments by training policies to map visual observations to actions from human demonstrations. However, these demonstrations are expensive to collect and modern game-playing is often done through streaming in which network delay and compression introduce spatiotemporally correlated visual artifacts that can cause a covariance shift at test time. To address these challenges, we propose streaming augmentations that mimic four types of artifacts commonly encountered during streaming with low-bandwidth network connection: pixelated blocks and scrubs, global blur, and ghosting. We instantiate our approach on top of predictive inverse dynamics models (PIDM), which combine future-state conditioning with an inverse dynamics policy in a learned latent space, and evaluate the impact of our augmentations across three tasks in modern 3D video games. %
Under stable streaming conditions, agents trained with spatiotemporal augmentations achieve up to 41\% higher evaluation performance compared to agents trained without augmentations under an identical data budget. When network lag is introduced, agents trained with augmentations degrade by only 7.45\% vs  49.82\% of the original performance for agents trained only with the original data. These results clearly indicate that spatiotemporal augmentations tailored for the streaming setting are a simple yet powerful tool to train robust and efficient game-playing agents.
\end{abstract}


\section{Introduction}

Offline imitation learning (IL) aims to train policies that replicate expert behavior from demonstrations without access to reward or additional environment interactions, and has been applied across robotics~\cite{schaal1999humanoid,fang2019survey}, autonomous driving~\cite{pan2020agile}, and games~\cite{pearce2022csdm,pearce2023diffusion,schaefer2023visual}, as surveyed in~\cite{osa2018algorithmic}. In commercial game pipelines, IL is particularly appealing because it eliminates the need for task-specific reward design and costly online training infrastructure. Instead, policies can be trained purely offline from collected gameplay demonstrations, making it a logical fit to train game-playing agents at scale. %

\begin{figure}[t]
    \centering
    \includegraphics[width=\linewidth]{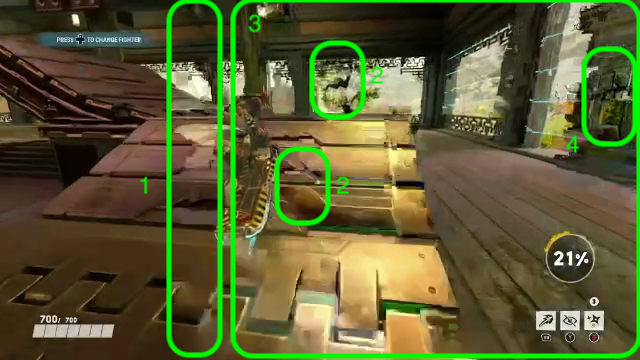}
    \caption{Streaming artifacts appearing under artificially induced network lag, simulating a low‑bandwidth user connection. The green numbered regions highlight common \emph{temporally correlated} artifacts that appear in bursts and persist across consecutive frames: \textbf{(1)} a vertical \emph{scrub/partial-frame corruption} band (a pixelated strip spanning the frame height), \textbf{(2)} \emph{localized macroblock pixelation} where compression blocks obscure scene details, \textbf{(3)} \emph{global fuzziness/blur} consistent with transient bitrate drops, and \textbf{(4)} \emph{ghosting/temporal smearing} where remnants of previous frames linger as patches or trails.}
    \label{fig:streaming_artifacts}
\end{figure}

In practice, however, two characteristics of the commercial game setting make offline IL particularly challenging. First, demonstration data is typically \emph{scarce}: each task requires dedicated collection by skilled human players, making large datasets costly and time-consuming to obtain. When trained on few demonstrations, agents are prone to overfitting and distribution shift during online rollouts leading agents to previously unseen states. Second, games are increasingly played in \emph{streaming} settings, where the game runs on a remote machine and video is transmitted through a streaming server. Network delay, bandwidth fluctuations, and compression introduce temporally correlated visual artifacts (see \Cref{fig:streaming_artifacts}) that might have been absent in the training dataset. Policies trained on data without such artifacts, especially in the small data regime, can fail catastrophically under these conditions, even when the underlying game dynamics are unchanged. %

Together, these challenges limit the efficacy of game-playing agents when deployed in streaming settings and motivate our guiding research question:
\emph{Can we train imitation learning agents that are robust to streaming artifacts from just a small number of demonstrations?}

We answer affirmatively by introducing \emph{streaming augmentations},\footnote{Code with more details available at \href{https://github.com/microsoft/temporal_streaming_augmentations_for_imitation_learning}{this Github repository}.} a new class of augmentations designed to model four types of visual artifacts that we commonly observe when streaming games under low-bandwidth conditions: fuzziness, pixelation, scrubs, and ghosting (see \Cref{fig:temporal_artifacts_examples}). We evaluate agents across three tasks in the two modern 3D video games and find that training agents with a combination of our streaming augmentations and standard image augmentations yields significant improvements in both sample efficiency and robustness under streaming conditions. When evaluated in streaming video games under injected lag, agents trained with our augmentations degrade by only 7.45\% of performance compared to evaluation under high-bandwidth network conditions, whereas non-augmented agents drop 49.82\%. Additionally, when evaluated in streamed games without injected lag, agents trained with our augmentations demonstrate significant sample efficiency gains, improving performance by up to 41\% compared to non-augmented agents trained with the same number of demonstrations.

\section{Related Work}

\textbf{Data augmentation for visual RL and IL.}
Augmentations are central to sample-efficient learning from pixels: RAD and DrQ showed that image augmentations can improve generalization and data efficiency in visual control~\cite{laskin2020rad,yarats2021drq}. Yadgaroff et al.\ studied augmentation for game-playing imitation learning agents~\cite{yadgaroff2024improvinggeneralizationgameagents}, CCIL introduced continuity-based augmentation for corrective imitation~\cite{ke2024ccilcontinuitybaseddataaugmentation}, and saliency-guided schemes have been shown to improve robustness under visual domain shifts in behavior cloning~\cite{pmlr-v270-zhuang25b}. Our work follows these ideas but targets a different failure mode: temporally correlated artifacts introduced by streamed video, motivating augmentations that are coherent over time rather than randomly sampled.

\textbf{Domain randomization and deployment robustness.}
Domain randomization aims to improve robustness to distribution shift by training models on data with randomized features (e.g., textures, lighting, camera)~\cite{tobin2017domainrandomization}. Beyond visual changes, streamed deployment introduces a channel shift: compression, lag, and buffering produce structured artifacts without changing underlying dynamics. Our streaming augmentations can be viewed as channel-focused randomization that perturbs the observation stream to match streamed gameplay conditions, complementing standard image augmentations.

\section{Background}
\label{sec:background}

\textbf{PIDM.}
We instantiate our approach on top of predictive inverse dynamics models (PIDM)~\cite{du2023textguideddp,xie2025latentdiffusion,tian2025pidm,anonymous2026pidm}, which improve data efficiency in offline IL by decomposing decision-making into predicting a future state and predicting the action to go from the current state to the predicted future state. 
In particular, we follow the implementation of \cite{anonymous2026pidm}. States are approximated with a history of recent video frames $h_t$ and the inverse dynamics model (IDM) policy outputs controller actions to represent continuous joystick controls and discrete button events. A key component is the visual representation: we use a \emph{frozen} pre-trained visual backbone from \textsc{Theia}~\cite{shang2024theia} to encode frames into latent features, and train a lightweight MLP for the IDM to predict actions from these features. 

\textbf{Standard image augmentations.}
We consider training PIDM models with standard image augmentations that are applied independently to each frame with the goal of improving robustness and sample efficiency of image-based agents. These augmentations increase the visual diversity of the dataset while preserving task semantics and the underlying game dynamics. We consider two types of augmentations commonly used to train game-playing agents from images~\cite{laskin2020rad,yarats2021drq,vpt}:
\begin{itemize}
    \item \textbf{Random affine transforms.} Small geometric perturbations (rotation, translation, scale, shear) that encourage invariance to minor viewpoint changes and camera jitter.%
    \item \textbf{Color jitter.} Photometric perturbations model variation in exposure and color grading. %
\end{itemize}

\textbf{Offline Training with Augmented Data}
\label{sec:offline_aug_pipeline}
Applying augmentations during training can slow-down training. To avoid this cost, we follow the setup from \cite{schaefer2023visual,anonymous2026pidm} and pre-compute and cache augmented variants of demonstrations. For each demonstration $\tau$, we generate $M$ augmented variants for a total of $(M{+}1)$ versions of each demonstration: the original version and $M$ augmented variants. Pre-computing the augmentations improves reproducibility by fixing the dataset and enabling deterministic re-runs as well as controlled ablations where different methods can use identical augmented variants. However, unlike on-the-fly augmentations, caching pre-computed augmented variants increases required storage, and limits the diversity of augmentations. During training, we sample the original (clean) demonstration with probability $p_{\mathrm{orig}}$ and sample any augmented variant with uniform probability to anchor the training distribution in the original visual frames while exposing the model to diverse visual corruptions.

\begin{figure*}[t]
    \centering
    \includegraphics[width=\textwidth]{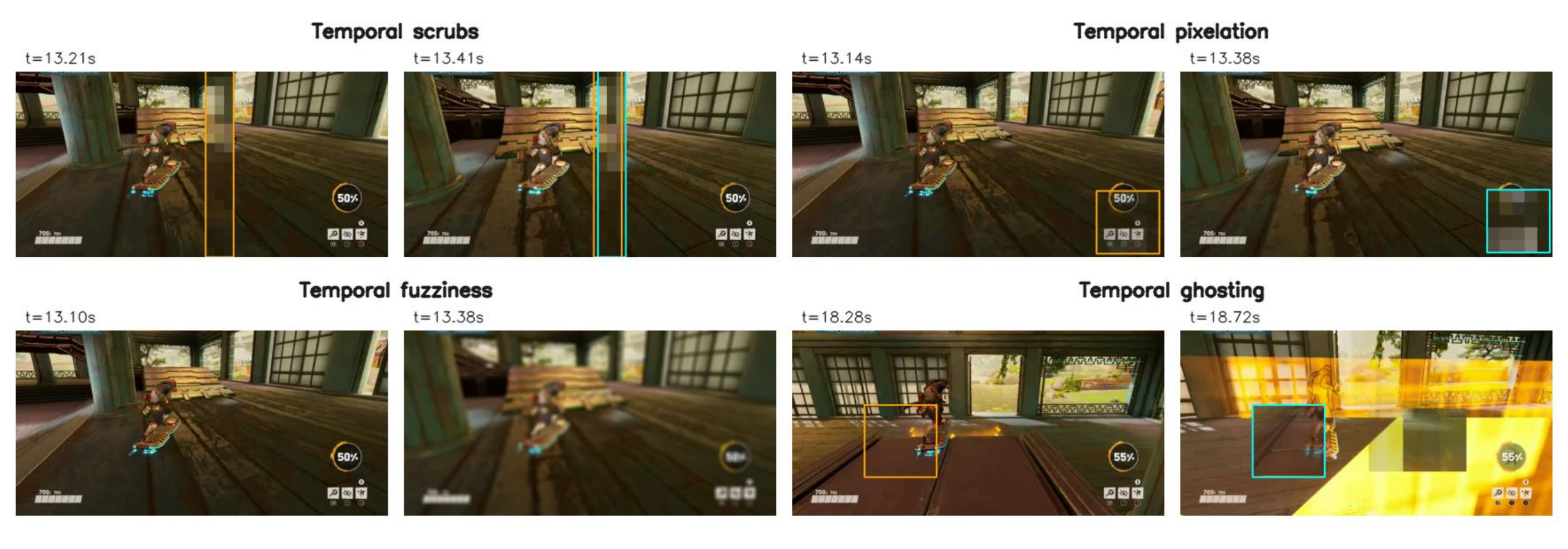}
        \caption{Examples of our streaming perturbations applied to two temporally close video frames.
        Colored boxes indicate the affected region at each timestamp.
        For \emph{scrubs}, the artifact is a thin vertical corrupted band (\textcolor{orange}{orange} band in the left figure) whose \emph{position} shifts over time (\textcolor{cyan}{cyan} band in the right); 
        For \emph{pixelation}, a localized region \textcolor{orange}{orange} box is down- and then up-sampled into blocky macroblocks and the pixelated region \textcolor{cyan}{cyan} box moves smoothly.
        For \emph{fuzziness}, a global blur is applied to the whole frame with smoothly increases in strength from the first to second frame. 
        For \emph{ghosting}, a region from the previous frame (\textcolor{orange}{orange} box) is blended into a new frame (\textcolor{cyan}{cyan} box), producing transparent overlays.}
    \label{fig:temporal_artifacts_examples}
\end{figure*}

\section{Streaming Augmentations}
\label{sec:method_temp_augs}

Visual artifacts under streaming conditions are \emph{temporally} and \emph{spatially correlated} and \emph{structured}. These artifacts typically persist for multiple frames, appear in bursts, and evolve smoothly as network conditions fluctuate. To train game-playing agents that are robust to these artifacts, we propose \emph{streaming augmentations}. In this section, we first introduce four hand-designed spatiotemporal perturbations that together compose our streaming augmentations.
Second, we describe how we apply these perturbations to model the smooth but bursty nature of visual artifacts.

\begin{figure}[ht]
    \centering
    \includegraphics[width=0.75\linewidth]{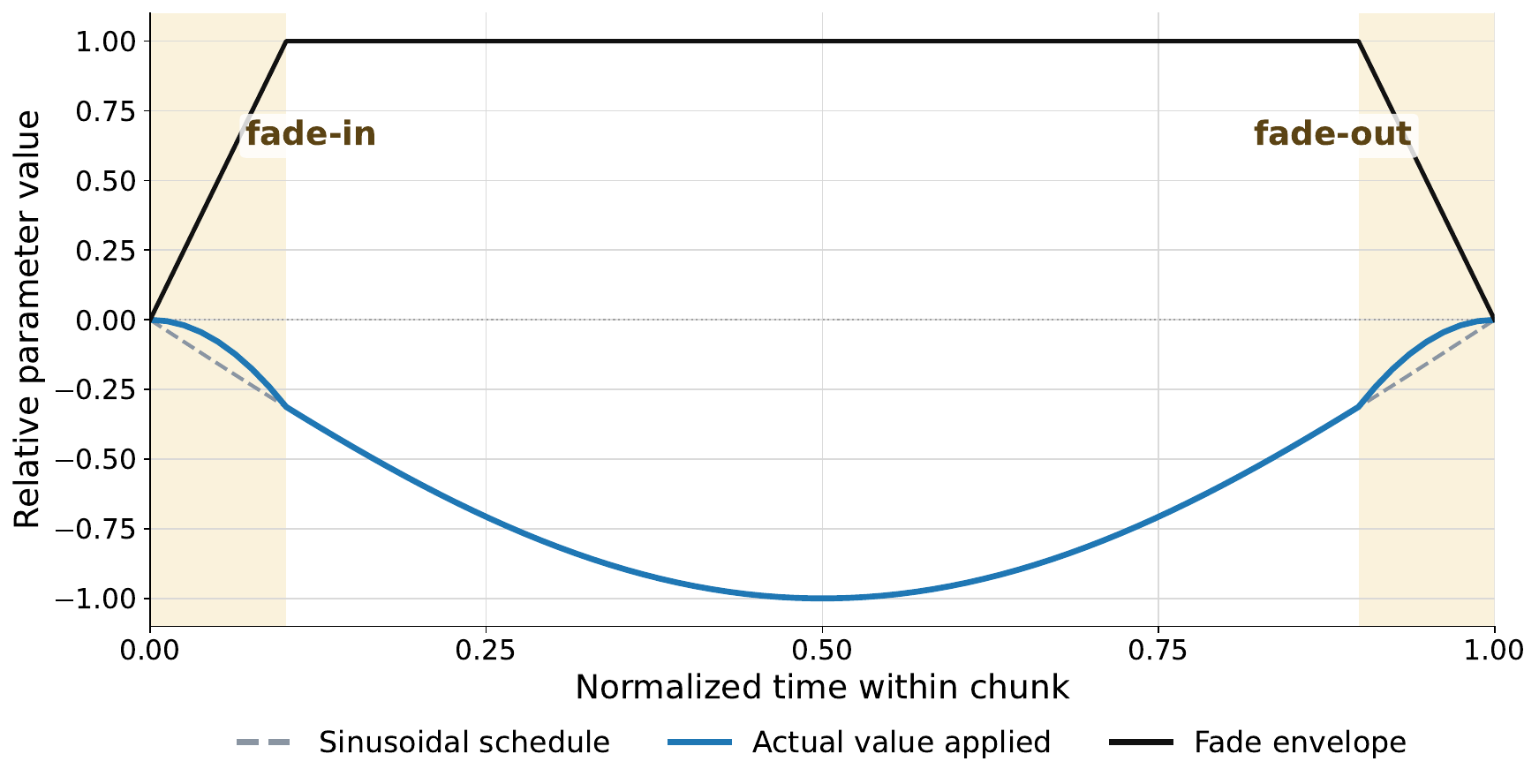}
    \caption{Exemplary modulation of a streaming perturbation parameter across normalized time within a chunk. We visualize the sinusoidal schedule, the fade-in and fade-out factor that suppresses perturbations near the beginning and end of a chunk, and the \emph{actual parameter value} obtained by multiplying the sinusoidal schedule with the fade factor. For readability, the fade-in and fade-out are slightly exaggerated compared to our experiments. We apply this scheduling to control parameters of streaming augmentations, such as the movement or intensity of perturbations.}
    \label{fig:temporal_augmentation_schedule}
    \vspace{-0.5cm}
\end{figure}

\begin{figure*}[t]
\centering
\includegraphics[width=\textwidth, height=10.7cm]{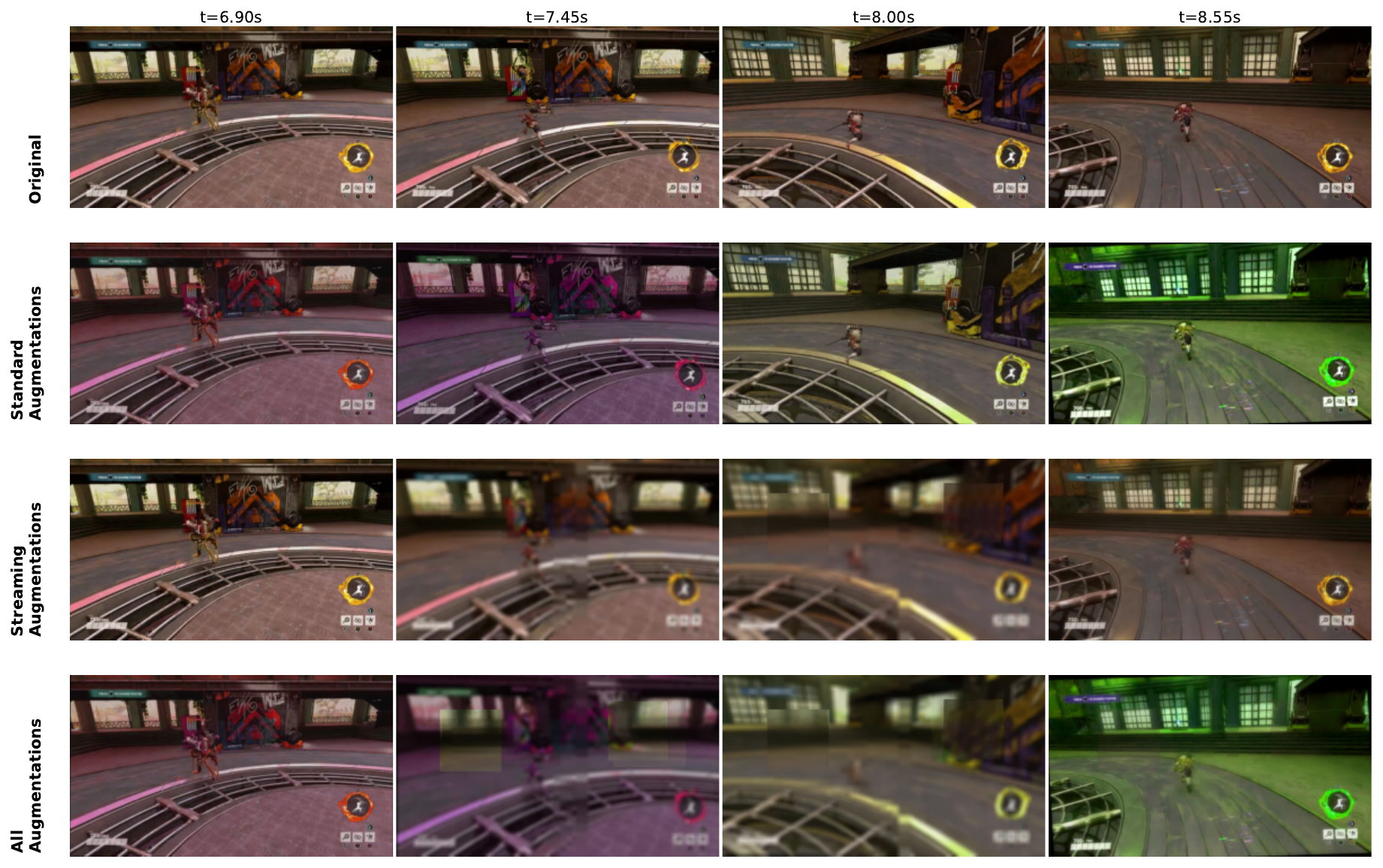}
\caption{Visualization of augmentations across four frames spanning $t \in [6.90\mathrm{s}, 8.59\mathrm{s}]$ of a demonstration. We show the original (i.e., non-augmented) video frames, frames with standard image augmentations, frames with our streaming augmentations, and frames with a combination of standard and streaming augmentations.}
\label{fig:aug_grid_gong}
\end{figure*}

\subsection{Spatiotemporal Streaming Perturbations}
\label{sec:method_perturbations}
\Cref{fig:temporal_artifacts_examples} visualizes each perturbation and their temporal modulation for representative consecutive frames. We proceed to explain each perturbation and their key parameters:

\textbf{Scrubs} produce one or multiple moving narrow vertical bands of pixelation that persist across consecutive frames and smoothly move throughout consecutive frames, resembling partial-frame corruption that can occur during streaming. To obtain the pixelation effect, we downsample the region of the image before upsampling it again to the original resolution, leading to a loss in details. Scrubs are parameterised by the width of their band, the pixelation factor that determines how aggressively the band is downsampled, and other parameters that determine the initial position and motion of the scrub.

\textbf{Pixelation} produces one or multiple moving squares of blocky regions consistent with compression artifacts that appear under bitrate drops and evolve as the stream quality fluctuates across frames. Similar to scrubs, the effect is achieved by down- and upsampling the region of each block within the image. Pixelation is parameterized by the size of the squared block, the pixelation factor, and parameters that determine the initial position and motion of the block. In contrast to scrubs, pixelation blocks move horizontally and vertically so we independently obtain parameters for both dimensions.

\textbf{Fuzziness} applies a Gaussian blur to the video frame consistent with global compression fallback during congestion, that fades in and out to resemble how streaming quality typically degrades and recovers. It is parameterized by the maximum strength of the blur, and parameters modulating how quickly the blur changes its strength over time. 
If the blur strength is less than a certain threshold, the blur is skipped for that frame.

\textbf{Ghosting} blends squared regions from previous (potentially already augmented) frames into the current frame, producing faint residual copies of objects by blending content from previous frames into the current frame, resembling inter-frame prediction errors or stale reference-frame artifacts in compressed video. Ghosting is parameterized by the size of its region, the opacity of the previous frame content \(\alpha\), and parameters determining the horizontal and vertical positions and motion of the perturbation (similar to pixelation blocks). 
The ghosting effect is defined as follows:
\[
\hat{x}_t[R_t] = (1-\alpha)x_t[R_t] + \alpha \hat{x}_{t-1}[R_t],
\]
where \(x_t\) denotes the current frame, \(\hat{x}_{t-1}\) the previous augmented frame, $\alpha$ the opacity, and $R_t$ the region of the effect. We emphasize that ghosting only copies regions from the single previous frame but since this frame might already contain regions with ghosting perturbation, the effect propagates recursively and allows residual content to persist across sequences of consecutive frames.

\subsection{Temporal and Spatial Modulation of Perturbations}
\label{sec:method_modulation}
In this subsection, we explain how we modulate the parameters of the perturbations to ensure smooth spatiotemporal augmentations.
We apply each perturbation to contiguous \emph{chunks} of frames rather than to individual frames independently, so perturbations persist for a sequence of frames stretching from fractions of a second to several seconds. A chunk $C$ is a sequence of \(T\) consecutive frames:
\(
C = (x_0, \dots, x_{T-1}),
\)
where \(T\) is sampled uniformly from a chunk-size range \([50,100]\). For each perturbation, we sample their parameters to determine factors such as the strength, position, and motion of the perturbation throughout the frames of the chunk. 
Throughout a chunk of \(T\) frames indexed by \(t \in \{0,\dots,T-1\}\), the parameters of perturbations follow (periodic) sinusoidal functions of the normalized time, $u_t$, given by:
\[
u_t = \frac{t}{\max(T-1,1)}.
\]
Here, \(t\) is the frame index within the chunk, and \(u_t \in [0,1]\) is the normalized position of the frame within the chunk. This modulation scheme ensures that perturbations evolve smoothly throughout the chunk instead of randomly flickering from frame to frame.

For example, for perturbations that move spatially throughout the chunk such as scrubbing, pixelation and ghosting, we parameterize their location over time as
\[
p_t = \operatorname{clip}\!\left(c + a \sin(2\pi \omega u_t + \phi)\right),
\]
with \(c\) denoting the anchor position of the perturbation within the chunk; \(a\) is the amplitude, which defines how far the perturbation can move from the anchor position; \(\omega\) is the frequency, which controls how quickly the perturbation moves; and \(\phi\) is the phase, which specifies the starting position within the motion cycle relative to the anchor. This sinusoidal modulation makes motions smooth and reversible rather than abrupt. We independently sample the parameters for the horizontal and vertical directions of each perturbation to allow for variable motions across the chunk. The \(\operatorname{clip}(\cdot)\) operator enforces image bounds so that the corrupted region never leaves the frame.

To ensure continuity across chunks and to avoid sudden visual changes between frames at the transition of two chunks, we define a fading mechanism to each chunk. The fading mechanism blends the augmented frame \(\tilde{x}_t\) with the original unperturbed frame, $x_t$, and resulting in a new frame, $y_t$:
\[
y_t = (1-\lambda_t)x_t + \lambda_t \tilde{x}_t,
\]
where the fade factor \(\lambda_t \in [0,1]\) controls the strength of perturbations at chunk time \(t\). \Cref{fig:temporal_augmentation_schedule} illustrates this modulation scheme and how we combine the sinusoidal parameter schedule with the fade envelope to produce the per-frame modulation of augmentations applied within each chunk. For our streaming augmentations, we use \textbf{3 fade frames} at the beginning and end of sufficiently long chunks, so perturbations fade in at the beginning of each chunk, remain strongest in the middle, and then fade out.

\begin{figure*}[t]
\centering
\includegraphics[width=\linewidth]{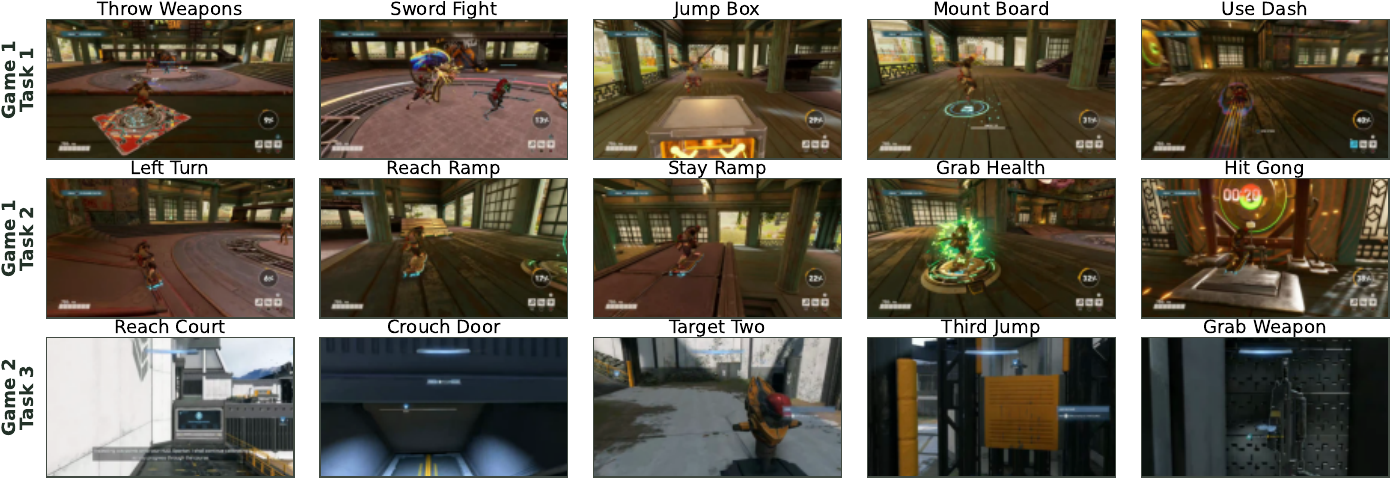}
\caption{Representative frames for key task milestones in each video game task.}
\label{fig:milestones_snapshot}
\end{figure*}

\section{Experimental Setup}
\label{sec:exp_setup}

All experiments follow the same training and evaluation protocol. We train agents purely offline from a fixed set of human demonstrations using a PIDM implementation similar to~\cite{anonymous2026pidm}: a frozen pretrained visual backbone~\cite{shang2024theia} with a small trainable state-encoder that produces latents $z_t$ of visual observations at time $t$; an MLP inverse-dynamics model (IDM) that predicts controller actions conditioned on $(z_t, z_{t+k})$, and a state predictor instantiated as a time-indexed lookup along a fixed demonstration that provides $z_{t+k}$. For our experimental setup we use the same offline training pipeline as described in \Cref{sec:offline_aug_pipeline}. For our experiments we use $M{=}10$ cached augmentations, as it strikes a good balance, yielding strong efficiency and robustness gains (see \Cref{tab:dojo_tour_real_streaming_noise} and \Cref{fig:sample_efficiency_dojo}) while retaining the diversity and keeping the cached dataset manageable. We also set the probability of sampling the clean trajectory $p_{\mathrm{orig}} = 0.2$.

After training, agents are evaluated online by deploying them in the streaming environment.
Across all experiments, we load the trained agent on a remote virtual machine (VM) to run inference and communicate predicted actions through a commercially available service to stream video games. 
During evaluation, the agent interacts with the environment at 30\,Hz, i.e.\ we require agents with an inference time smaller than $33$\,ms. At each step, the agent receives the current video frame, predicts a controller action, and sends the predicted action to the game environment through the streaming client.
The game itself runs on a remote server in a different cloud region, and the distance between the VM and the game server induces bidirectional random latency in the video stream and controller actions, providing a source of channel noise for our experiments. We keep the underlying game content and task setup fixed across evaluations. For the robustness experiments, we study the distribution shift induced by visual artifacts in the received video frames.

\subsection{Baselines and Approaches}
During training, we use four training datasets (\Cref{fig:aug_grid_gong}): 
\begin{itemize}
    \item \textbf{No augmentations.} Original demonstrations only ($M{=}0$).
    \item \textbf{Standard augmentations.} Combination of the original demonstrations and demonstrations with standard augmentations ($M=10$).
    \item \textbf{Streaming augmentations (Ours).} Combination of the original demonstrations and demonstrations with our streaming augmentations ($M=10$).
    \item \textbf{All augmentations (Ours).} Combination of the original demonstrations and demonstrations augmented with standard and our streaming augmentations. We precompute 5 standard and 5 streaming augmentations for a total of $M=10$ precomputed augmentations.
\end{itemize}

\subsection{Tasks and video games}

We evaluate on three tasks across two video games:
\begin{itemize}
    \item \textbf{Game 1.} We define two tasks: \emph{Task 1} and \emph{Task 2}, consisting of 35 to 45 seconds of navigation and object interactions.
    \item \textbf{Game 2.} We define a multi-stage long-horizon task (Task 3) in Game 2 requiring about 2:32 minutes of navigation, and precise discrete action timings to crouch, jump, and interact with objects. Taking an action even just a few steps too early or late can result in failure of any subsequent objectives within the mission. 
\end{itemize}
For each task, we collect a dataset of 30 demonstrations of human expert gameplay for training.

\subsection{Evaluation method and metrics}
\label{sec:metrics}

We evaluate each agent's performance based on its ability to complete predefined milestones for each task, like successful navigation checkpoints, jumps, crouch actions, object interactions and attacks (\Cref{fig:milestones_snapshot} shows five key milestones of each task). We mark a rollout as completing a milestone if the agent completes the milestone event \emph{at any time} during the episode, i.e.\ the agent is not required to precisely follow the demonstration timings. This choice reflects the goal of measuring whether the agent achieves the intended behavior, rather than reproducing the exact demonstration.

To reliably determine whether agents complete each milestone during evaluation rollouts, we manually annotate all rollout videos. To minimize annotation bias, videos are anonymized and randomly sampled so that annotators are unaware of which method produced each rollout.

We report the aggregate percentage of milestones completed by each method as main metric. For a more granular view, we additionally present cumulative per‑milestone completion rates, which reveal how consistently each agent succeeds at individual milestones throughout the task.

Unless mentioned otherwise, we train five seeds for each configuration, evaluate the model of each seed for ten rollouts, and report aggregate metrics across seeds. We compute each aggregate metric by first averaging the metric across rollouts for each seed, and then reporting the mean and standard error computed over seeds.

\section{Results}
\subsection{Sample Efficiency Gains}
\begin{figure}[t]
\centering
\includegraphics[width=\linewidth]{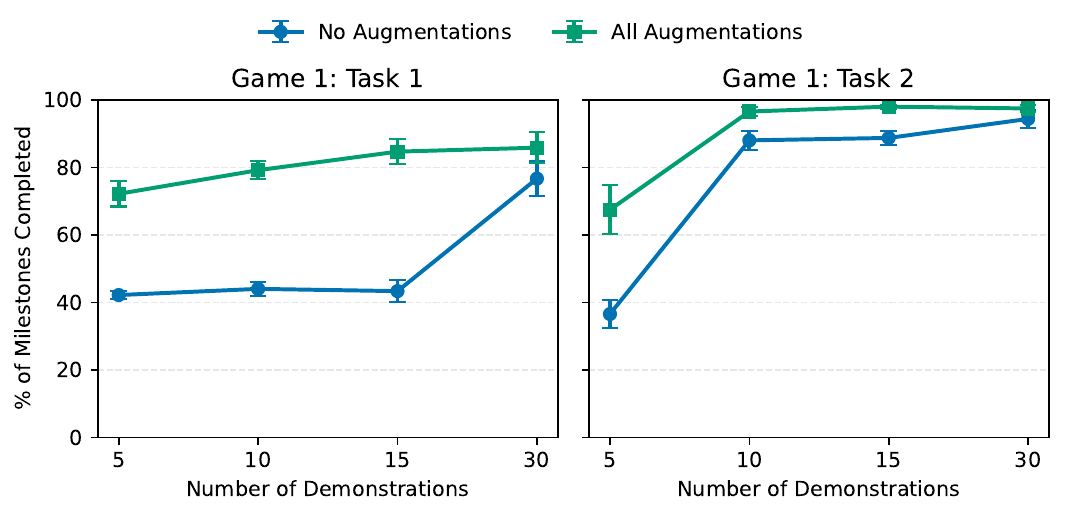}
\caption{
Milestone-based sample-efficiency results in Task 1 and Task 2 of Game 1. 
}
\label{fig:sample_efficiency_dojo}
\end{figure}
\Cref{fig:sample_efficiency_dojo} shows the sample efficiency of PIDM agents trained with or without all augmentations on tasks of Game 1. Across both tasks, we find that PIDM trained with all augmentations consistently achieve higher milestone completion than agents trained without augmentations, as expected. Performance gains are largest in the lowest data regime where agents are trained on just five human demonstrations. In this case, agents trained with augmentations reach up to 42\% and 31\% higher milestone completion rate in Task 1 and Task 2, respectively, compared to agents only trained on the original demonstrations.

In Task 1, the benefits from augmentation are substantial at 5--15 training demonstrations: augmented PIDM reaches roughly the 70--85\% milestone-completion range, while non-augmented PIDM remains near 40--45\% and often fails to progress to later milestones. Similarly, in Task 2, augmented PIDM starts at a higher performance for five training demonstrations and reaches near-perfect performance of $\sim95\%$ milestones completed with just ten training demonstrations. In contrast, non-augmented PIDM requires 30 training demonstrations to approach similar performance. %

\subsection{Ablating Augmentation for Sample Efficiency Gains}
\label{sec:exp_sample_eff_ablation}
\begin{figure}[t]
    \centering
    \includegraphics[width=\linewidth]{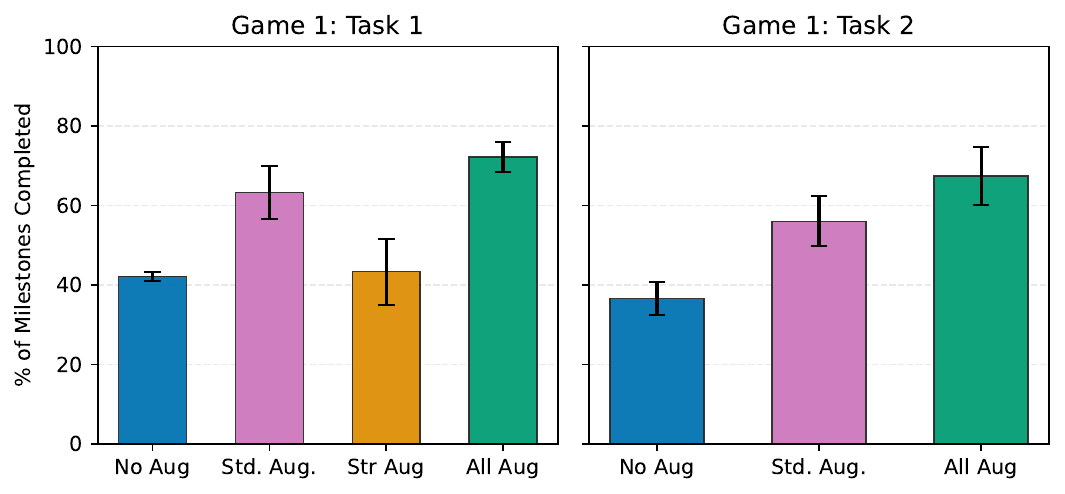}
    \caption{Milestone completion for agents trained on \textbf{5} demonstrations in Task 1 and Task 2 of Game 1.}
    \label{fig:sample_efficiency_aug_types_5}
\end{figure}
\begin{figure}[t]
    \centering
    \includegraphics[width=\linewidth]{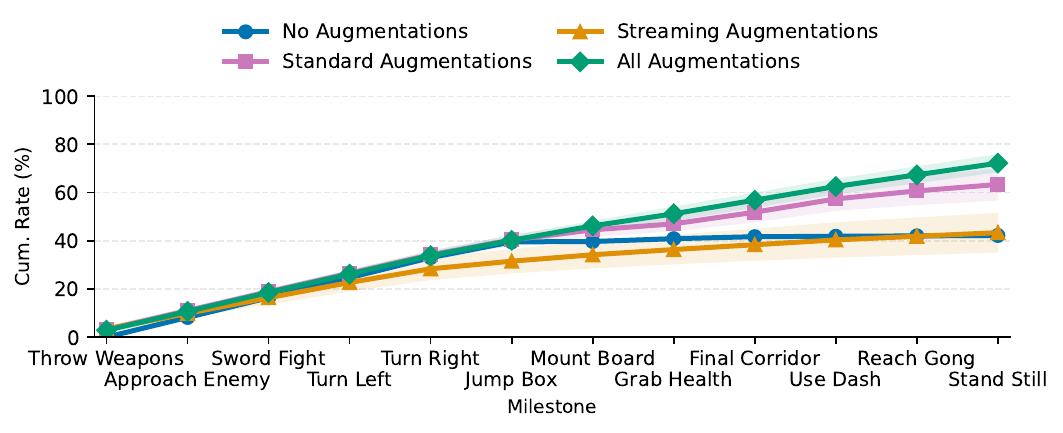}
    \caption{Cumulative milestone completion rate in Task 1 of Game 1 under normal network conditions for agents trained with 5 demonstrations. Each point shows the running completion performance aggregated from the first milestone up to that position, normalized by the total number of milestones.}
    \label{fig:ex_5_samples_milestone_breakdown}
\end{figure}

\Cref{fig:sample_efficiency_dojo} shows that augmentations provide significant sample efficiency gains. But \emph{which components are responsible for these gains?} To answer this, we ablate the augmentation pipeline for the lowest data regime with five training demonstrations where performance gains were the most pronounced and compare agents trained with different augmentation subsets on both tasks of Game 1.

As shown in \Cref{fig:sample_efficiency_aug_types_5}, standard image augmentations already improve milestone completion over the no-augmentation baseline, consistent with prior work~\cite{yadgaroff2024improvinggeneralizationgameagents, ke2024ccilcontinuitybaseddataaugmentation, pmlr-v270-zhuang25b} showing that frame-wise visual augmentations can improve sample efficiency in imitation learning from pixels. However, combining these standard image augmentations with our proposed spatiotemporal streaming augmentations further improves performance, indicating that our streaming augmentations provide additional benefits beyond standard image augmentation.

Interestingly, in Task 1, we find that only training agents with our streaming augmentations performs substantially worse than the combined setting. To localize this performance drop, \Cref{fig:ex_5_samples_milestone_breakdown} reports per milestone completion rates in Task 1 with 5 demonstrations. The combined pipeline improves completion on later milestones (e.g. ``Grab Health") indicating augmentation diversity improves the agent robustness. This suggests that our streaming augmentations are complementary with standard image augmentations: standard image augmentations introduce  variation in static images, while streaming augmentations expose the policy to structured spatiotemporal corruption patterns.
Together, these augmentations increase in diversity of the resulting training data, enabling agents to learn policies that are invariant to more types of visual distortions.

\begin{figure}[t]
    \centering
    \includegraphics[width=\linewidth]{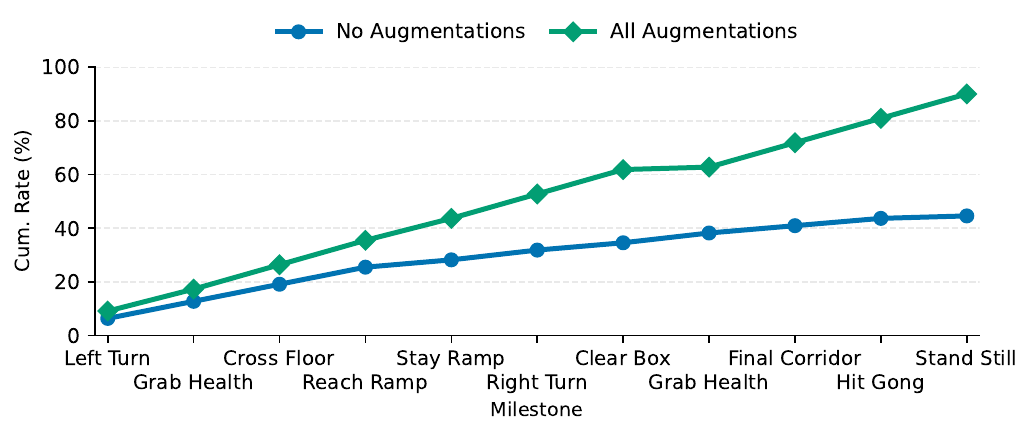}
    \caption{Cumulative per-milestone completion rate in Task 2 of Game 1 under induced streaming lag.}
    \label{fig:gong_real_streaming_milestone_breakdown}
\end{figure}
\subsection{Robustness to Streaming Artifacts}

\begin{table}[t]
\centering
\caption{Milestone completion percentages for agents trained on 30 demonstrations and evaluated in Task 2 of Game 1 with and without injected network lag. No-lag results report the mean percentage across five seeds, while lag results report the mean over ten rollout episodes under real streaming noise.}
\label{tab:dojo_tour_real_streaming_noise}
\begin{tabular}{lccc}
\toprule
Method & Regular (\%) & Injected Lag (\%) & $\Delta$ (\%) \\
\midrule
No Augmentations & 94.36 & 44.55 & 49.82 \\
All Augmentations & \textbf{97.45} & \textbf{90.00} & \textbf{7.45} \\
\bottomrule
\end{tabular}
\end{table}

We next evaluate whether our streaming augmentations improve the robustness of game‑playing agents to visual artifacts that occur under low‑bandwidth or high‑latency streaming conditions. To simulate these conditions, we inject additional network lag (2–10\,ms) during evaluation, increasing the likelihood that the video frames received by the agent contain of streaming‑induced visual artifacts.

\Cref{tab:dojo_tour_real_streaming_noise} summarizes the results. 
Under injected lag, agents trained with streaming augmentations lose just 7.45\% of their milestone‑completion performance compared to normal conditions, while non‑augmented agents lose 49.82\%, underscoring the significant robustness gains from our approach.

We emphasize that all PIDM agents use a visual backbone that has been pre-trained on large-scale image data with standard augmentations and is expected to already provide some robustness to visual variation. However, we find that such pre-trained visual backbones are insufficient to be invariant to the spatiotemporal distortions induced during streaming. \Cref{fig:gong_real_streaming_milestone_breakdown} further localizes these failures. Under lag, the non-augmented agents often fail to complete later milestones (e.g., ramp traversal and final gong), whereas agents trained with our augmentations maintain high completion rates even in later milestones.

\subsection{Robustness under Synthetic Streaming Corruption}

To verify whether the standard and streaming augmentations interact synergistically, as opposed to conflicting with each other, we evaluate agents under a controlled synthetic stress test. 
In this experiment, we train agents on 30 demonstrations and, during evaluation, apply our streaming augmentations to corrupt $\sim$50\% of video frames in each episode. This setup creates a severe but reproducible observation‑channel mismatch, enabling a direct comparison across augmentation methods under identical corruption patterns.

\begin{figure}[t]
    \centering
    \includegraphics[width=\linewidth]{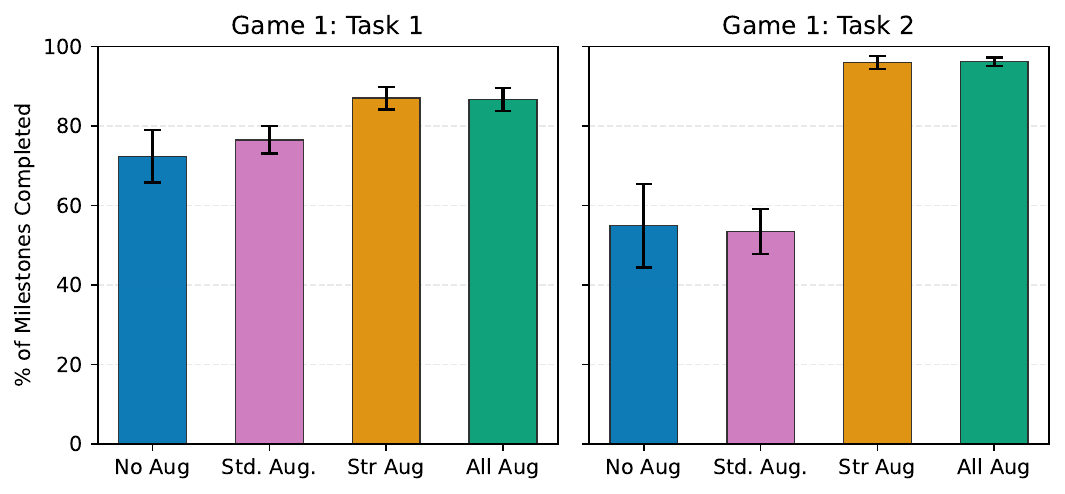}
    \caption{Milestone completion for agents trained on \textbf{30} demonstrations and evaluated under streaming augmentations in both tasks of Game 1.}
    \label{fig:robustness_synthetic_aug_types}
\end{figure}

\Cref{fig:robustness_synthetic_aug_types} shows that standard image augmentations yield only marginal robustness gains over the no‑augmentation baseline, confirming that they are insufficient to address spatiotemporal distortions. As expected, agents trained with streaming augmentations maintain high milestone‑completion rates, as they are invariant to these distortions: Task 1 87\% vs. 76.5\% for standard augmentations, and Task 2 96.18\% vs. 53.45\%. 

Finally, training agents with both standard and streaming augmentations performs on par with those trained with streaming‑only augmentations. This confirms that the two are complementary and combining them preserves the robustness benefits of streaming augmentations while introducing additional diversity during training, an effect that also improves sample efficiency, as shown in \Cref{sec:exp_sample_eff_ablation}.

\subsection{Generalization to other games}

\begin{table}[b]
\centering
\caption{Milestone completion for agents trained on 30 demonstrations and evaluated in Task 3 of Game 2.}
\label{tab:game2}
\begin{tabular}{lc}
\toprule
Method & Milestones Completed (\%) \\
\midrule
No Augmentations & 55.20 $\pm$ 4.74 \\
All Augmentations & \textbf{65.07 $\pm$ 4.83} \\
\midrule
Gain & +9.87 \\
\bottomrule
\end{tabular}
\end{table}

The results so far show that our augmentations improve both sample efficiency and robustness on two tasks within \emph{Game 1}.
A natural question is whether these benefits are specific to these tasks and this game, or whether they transfer, without modification, to other games with different dynamics and visual characteristics.

To test this, we evaluate in Game 2 on the Task 3. 
This setting differs from the Game 1 tasks in several important ways: it has a substantially longer horizon, requires precise timing of discrete actions (e.g., crouching, jumping, and scripted interactions), and presents a distinct visual style. 
In this experiment, all agents are trained on 30 demonstrations.

As shown in \Cref{tab:game2}, PIDM trained with augmentations completes $10\%$ more milestones within Task 3 of Game 2 than the non-augmented PIDM baseline. This indicates that our augmentations are not specific to Game 1 but can provide benefits in other video game environments though the magnitude of these improvements vary across games and tasks. 

\begin{figure}[t]
    \centering
    \includegraphics[width=\linewidth]{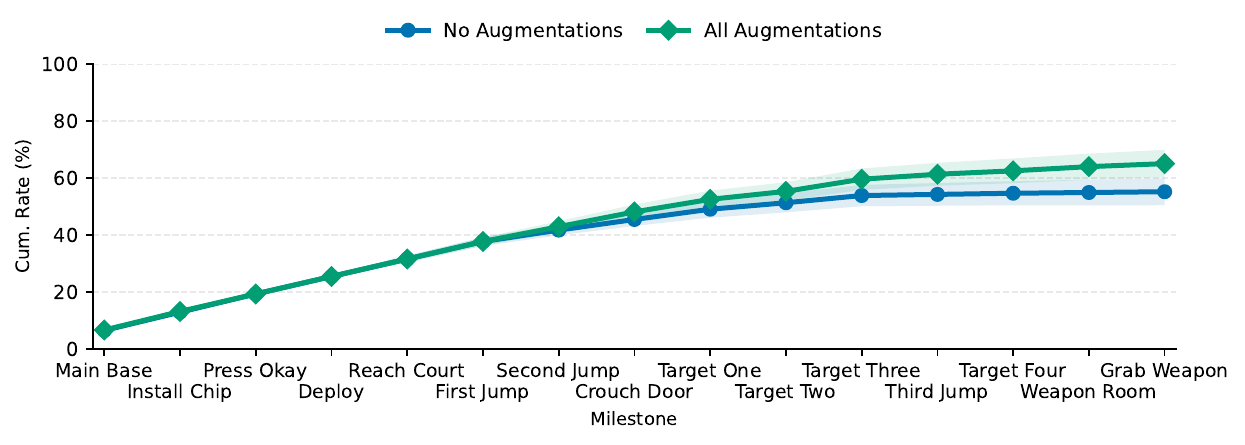}
    \caption{Cumulative per-milestone completion rate in Task 3 of Game 2 under normal network conditions.}
    \label{fig:game2_milestone_breakdown}
\end{figure}

Beyond improving low‑level action execution, our augmentations also enhance stage‑to‑stage reliability in long‑horizon settings, where an early failure can prevent the agent from reaching any later milestones. To understand where these benefits arise, \Cref{fig:game2_milestone_breakdown} reports per‑milestone completion rates. Overall, completion rates tend to decline over the course of the task, since failing a key milestone leaves the agent “stuck” and unable to progress further. This compounding effect is most visible at bottleneck milestones, where we observe sharp drops in completion: ``Second Jump" and ``Third jump". For the ``Third Jump" milestone in particular, we see significant gains from our augmentations, with non-augmented PIDM agents only succeeding at this milestone in 6\% of rollouts, while agents trained with our augmentations succeed 26\% of the time. Furthermore, our augmentations improve not only completion rates for early milestones but also completion deeper into the tutorial, where failures are typically caused by compounding errors in movement, timing, and interactions.

\section{Discussion and Conclusion}
We propose plug-and-play streaming augmentations that mimic spatiotemporally correlated distortions encountered under streaming conditions, and combine them with standard image augmentations into a practical training pipeline for imitation learning agents. When paired with PIDM agents and evaluated in three tasks across two commercial 3D games, our augmentations yield consistent gains in sample efficiency and streaming robustness. 
In the small-data regime, augmented agents achieve up to 41\% higher milestone completion than non-augmented agents, and under injected network lag they drop only \(7.45\%\) of their performance under high bandwidth network conditions compared to \(49.82\%\) for agents trained only on original non-augmented data. These improvements persist at action-critical bottlenecks and generalize to the long-horizon Task 3.

Future work could evaluate our proposed streaming augmentations across a broader range of streamed games and network conditions to understand their impact in varying conditions. It would also be valuable to explore online adaption of the augmentation strength based on measured stream quality, rather than using a fixed schedule. Finally, future work could study how streaming augmentations interact with different pretrained visual encoders and imitation learning objectives.

\bibliographystyle{IEEEtran}
\bibliography{ref}
\end{document}